# Evaluating Fair Feature Selection in Machine Learning for Healthcare

Md Rahat Shahriar Zawad[1,2], Peter Washington[1,2], PhD
[1]Department of Information and Computer Sciences, University of Hawaii at Manoa, Honolulu, HI, USA.
[2]Hawai'i Digital Health Laboratory, Honolulu, HI, USA.

**Abstract**
*With the universal adoption of machine learning in healthcare, the potential for the automation of societal biases to further exacerbate health disparities poses a significant risk. We explore algorithmic fairness from the perspective of feature selection. Traditional feature selection methods identify features for better decision making by removing resource-intensive, correlated, or non-relevant features but overlook how these factors may differ across subgroups. To counter these issues, we evaluate a fair feature selection method that considers equal importance to all demographic groups. We jointly considered a fairness metric and an error metric within the feature selection process to ensure a balance between minimizing both bias and global classification error. We tested our approach on three publicly available healthcare datasets. On all three datasets, we observed improvements in fairness metrics coupled with a minimal degradation of balanced accuracy. Our approach addresses both distributive and procedural fairness within the fair machine learning context.*

**Introduction**
As machine learning becomes increasingly considered for clinical decision support for a broad spectrum of conditions such as cancer[1–3], cardiovascular diseases[4,5], mental health disorders[6–8], and infectious diseases[9,10], algorithmic biases continue to be observed based on attributes such as skin color and gender[11–13]. This disparity in performance often stems from historical socioeconomic and cultural biases[14], raising concerns about the adequacy of machine learning models to serve all patient groups.

A growing body of research has been conducted to mitigate bias in machine learning models, particularly through optimization and regularization[15–17], recalibrating models[18,19] and preprocessing of data[20–22]. However, research into fairness during the feature selection process, a common step in the classical machine learning pipeline, is relatively sparse by comparison. While prior studies highlight the necessity of fairness-aware frameworks in feature selection, they frequently overlook the complexities associated with the subtle interactions among differing definitions of fairness[23,24]. Additionally, investigations into information-theoretic methods and kernel alignment have failed to adequately consider the impact of demographic-specific feature dependencies[25,26]. Although there are proposals to assess procedural fairness through societal perceptions and legal standards[27–29] and distributive fairness using genetic algorithms by aiming to balance fairness and accuracy in machine learning predictions[25,26,30–32], a significant gap remains in creating an integrated, practical feature selection approach that accommodates varying definitions of fairness.

To address these issues, we introduce a generalized method for addressing biases in machine learning datasets and corresponding models during the feature selection process. We analyze three different health datasets, stratified by gender, to assess and select features separately per gender. By consolidating multiple feature selection methods into a unified framework and utilizing a combined metric for final feature selection, our approach addresses distributional fairness and offers a sophisticated perspective beyond conventional procedural fairness analyses. This methodological approach presents a novel means of achieving machine learning fairness that can be combined with other more well-known algorithmic fairness procedures to achieve a comprehensive strategy for reducing bias.

**Methods**

*Datasets*
We used three distinct healthcare datasets to evaluate the applicability and functionality of our approach across different scenarios (Table 1):
- Tappy Keystroke Dataset[33,34]: This dataset comprises keystroke timing information from 103 individuals, of which 32 have been diagnosed with mild Parkinson's Disease (PD) and 71 do not have PD. It is utilized to ascertain the presence of *PD* in subjects based on their keystroke dynamics.

- Clinical and Molecular Features data for Glioma Grading[35]: This dataset, derived from The Cancer Genome Atlas (TCGA) Project, includes data from 20 genes and 3 clinical attributes across 839 instances. It is used to differentiate between Lower-Grade Glioma (LGG) and Glioblastoma Multiforme (GBM).
- Hospital Admission Data for Coronary Artery Disease[36]: This is a processed version of the Hospital Outcomes dataset[37] collected from Hero Dayanand Medical College, Heart Institute Unit of Dayanand Medical College and Hospital. The dataset encompasses 53 features covering demographic information, admission specifics, laboratory results, and comorbidities, with a total of 6611 data points.

**Table 1. Summary of the datasets used for the study.**

| Dataset | # of Features | # of Datapoints | Protected Attribute | Prediction Task |
|---|---|---|---|---|
| Tappy Keystroke Dataset | 31 | 83 | Gender | Parkinson's Disease |
| Clinical and Molecular Features data for Glioma Grading | 22 | 839 | Gender | Clinical Glioma |
| Hospital Admission Data for Coronary Artery Disease | 53 | 6611 | Gender | Coronary Artery Disease |

*Preprocessing*

Among the three datasets utilized in this study, the Tappy Keystroke dataset was subject to the most extensive preprocessing. Erroneous entries were removed from the initial dataset, and metrics such as HoldTime and LatencyTime were standardized to ensure consistency. Subsequently, the refined data were categorized based on the participant's identity and the hand used for typing. Key statistical metrics—mean, standard deviation, skewness, and kurtosis—were then calculated for each user's keystroke durations and intervals[33]. This process of refinement and transformation resulted in the generation of 31 features for each user, and the cleaning procedure resulted in the exclusion of 20 users, leaving 83 instances for analysis.

In contrast, the Clinical and Molecular Features data for Glioma Grading and the Hospital Admission Data for Coronary Artery Disease underwent minimal preprocessing, as they were already in a substantially processed form. Enhancements were confined to the label encoding of categorical features to facilitate more effective model training and the elimination of features that were not relevant to the objectives of the study.

*Feature Ranking*

To assess feature significance in relation to gender discrepancies, we first divided the dataset according to the 'gender' attribute, thereby generating distinct subsets for male and female participants. Subsequently, various feature ranking techniques were applied independently to each data subset:
- Forward Feature Selection (FFS) with Random Forest (RF): The incremental contribution of each feature to an RF model's performance is assessed when iterative adding features[30].
- Recursive Feature Elimination (RFE) with RF: In contrast to FFS, RFE removes features iteratively[30].
- Decision Tree (DT) Depth: Features are evaluated based on their position in the DT. Features used at the root nodes are deemed more critical than those used at a deeper branch[38].
- Logistic Regression (LR) Weights: The absolute weights in an LR model signify the feature's effect on the outcome. Heavier weights suggest a stronger relationship with the dependent variable, offering a linear perspective on feature importance[39].
- Mutual Information (MI): Features with less mutual information suggests a weaker relationship, indicating a lower information released to the model about the gender[40].
- Spearman's Rank Correlation: Utilizing non-parametric measures, this method identifies features with lower correlation to the gender column, suggesting their diminished significance in gender-specific contexts[41].

After the individual ranks of the features were extracted for each demographic-specific dataset using each of these methods, the average ranking from these methods yielded the final feature ranking. The mathematical representation for each demographic subset ($R_{\text{final, demographic group}}$) combines the individual ranks from each method ($R_{\text{method, demographic group}}$), averaged over all utilized methods (N):

$$R_{\text{final,demographic}} = \frac{1}{N}\sum_{i=1}^{N} R_{\text{method } i, \text{demographic}}$$

An aggregated ranking, reflective of all demographic groups, is computed by averaging these demographic-specific ranks, providing a holistic view of feature significance. Here, 'M' represents the number of demographic groups:

$$R_{\text{final,combined}} = \frac{1}{M} \sum_{i=1}^{M} R_{\text{final,demographic group } i}$$

By dissecting and synthesizing feature importance through this multifaceted lens, the final ranking is less biased.

*Selection From Ranked Features*
We implemented Stratified K-Fold cross-validation with the value of K tailored to each dataset: we used 4-fold cross validation for the Tappy Keystroke dataset and 5-fold strategy to the other datasets. This differentiation accommodates the particular attributes and distributions of each dataset, ensuring the preservation of class proportions within each fold—an essential factor for reducing bias in model performance metrics[42].

The final feature selection was conducted iteratively, where the next features in the rank were added progressively and the chosen features were used to train a LR model. The efficacy of this model was assessed based on Balanced Accuracy (Bacc) and Disparate Impact (DI), with DI used as an indicator of fairness.

The combined metric, aimed at integrating both performance and fairness into a single optimization criterion, was computed as follows:

$$CombinedMetric = w_{acc} \times Bacc + w_{fair} \times (1 - |1 - DI|)$$

In this equation, $W_{(acc)}$ and $W_{(Fair)}$ represent the weights assigned to Bacc and fairness, which are set to 0.50. The optimal feature set following the ranking was selected based on the optimal value of the combined metric. This balanced approach allowed us to identify feature sets that optimize both prediction accuracy and fairness during the feature selection process, addressing potential biases and ensuring the model's decisions are equitable across different demographic groups.

*Evaluation Metrics*
We integrate four evaluation metrics to ensure both fairness and overall performance: Equalized Odds (EqO), Statistical Parity (SP), DI, and Bacc. EqO ensures that the True Positive Rate (TPR) and False Positive Rate (FPR) remain consistent across groups, defined as $P(\hat{Y} = 1|Y = y, G = g) = P(\hat{Y} = 1|Y = y, G = g')$ where $Y$ represents the actual outcome and $\hat{Y}$ resents the predicted outcome for different demographic groups $G$. SP aims to achieve equal positive prediction rates across different groups $g$ and $g'$, quantified as $P(\hat{Y} = 1|G = g) - P(\hat{Y} = 1|G = g') = 0$. DI assesses fairness through the ratio of positive outcomes, targeting a value of $\frac{P(\hat{Y} = 1|G = g)}{P(\hat{Y} = 1|G = g')} \approx 1$. Bacc calculates the mean of correct predictions for each class: $Bacc = \frac{1}{2}(TPR + TNR)$. These metrics collectively guide our model towards minimizing error while maximizing fairness[43–46].

*Experimental procedure*
We used correlation-based feature selection across all datasets. This initial step served as a benchmark for subsequent analyses. We then aimed to use the optimal subset of features that foster equitable outcomes. Specifically, for the Tappy Keystroke dataset, a 4-fold cross-validation approach was adopted, supplemented by 100 iterations of bootstrapped sampling within each fold to ensure the robustness and validity of our findings. The Glioma Grading and Coronary Artery Disease datasets were analyzed using a 5-fold cross-validation scheme, again with 100 bootstrapped samples for each fold. We used an LR model to evaluate the effectiveness of both the baseline and the novel fair feature selection methodologies. Figure 1 illustrates a summary of the data processing, feature selection, and evaluation strategy.

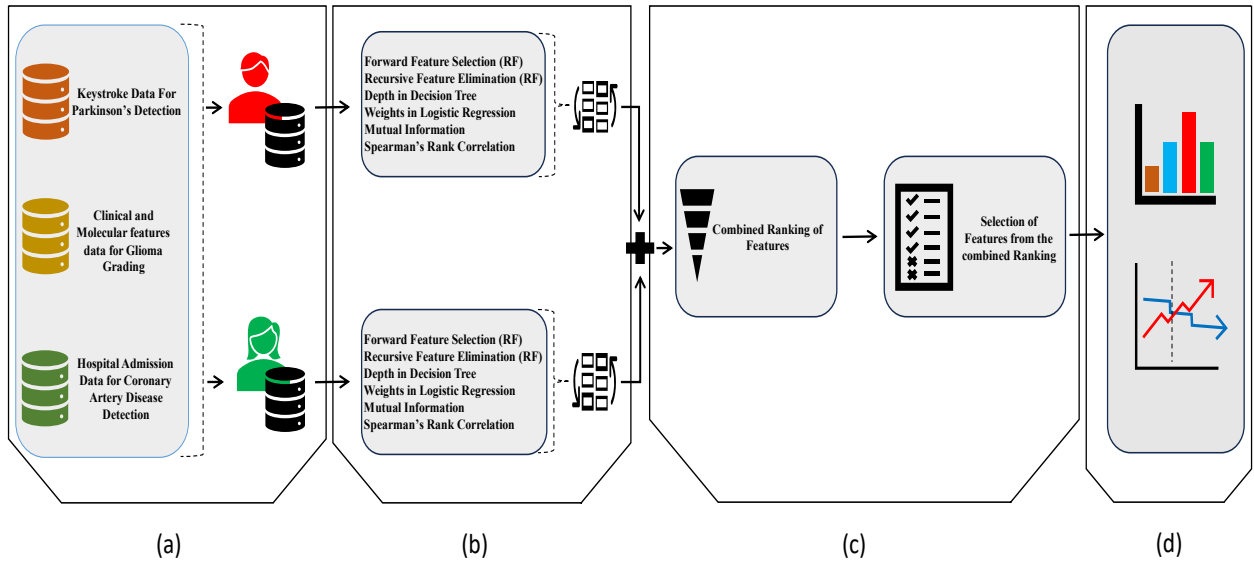

**Figure 1.** Overview of the feature selection and evaluation process: (a) Initial dataset preprocessing and stratification by gender to ensure equal representation and analysis. (b) Application of multiple feature ranking methodologies including FFS, RFE, DT Depth, LR Weights, MI, and Spearman's Rank Correlation to identify the most predictive features for each subset. (c) Integration of individual feature rankings into a unified, combined ranking to inform the final feature selection. (d) Comprehensive assessment of the resulting model's performance and fairness using established metrics, conducted across diverse datasets.

**Results**

*Tappy Keystroke Data*
For the Tappy Keystroke Dataset, fairness-oriented feature selection identified 19 features and we selected top 19 features from correlation-based feature selection for an equivalent baseline. Through the use of a 4-fold cross-validation combined with 100 rounds of bootstrapped sampling, we documented a notable improvement in fairness metrics: SP was refined to -0.0003 ± 0.0764 (from -0.0095 ± 0.1531), DI was reduced to 0.9822 ± 0.2127 (from 1.1846 ± 0.4026), and EqO saw an increase to 0.0542 ± 0.0638 (from -0.0667 ± 0.2424), indicating an increase in fairness across all metrics. We also observed a marginal decline in Bacc from 0.8262 ± 0.0578 to 0.7637 ± 0.0268. This tradeoff is depicted in Figure 2. Table 2 summarizes the results and presents the metrics used for comparison. Figure 3 demonstrates a comparison of fairness metrics and Bacc for standard and fair feature selection methods individually and as applied to the Tappy dataset.

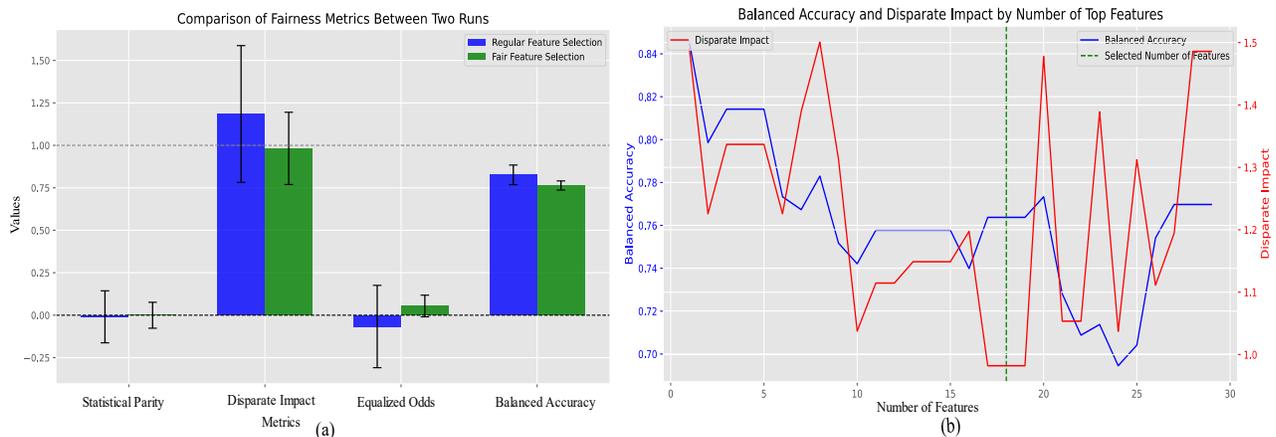

**Figure 2.** Tappy Keystroke dataset analysis: (a) Contrast in fairness metrics (DI, SP and EqO) and Bacc between conventional feature selection and fairness-oriented feature selection approaches. (b) The variations in fairness metrics and accuracy against the number of top features selected from the ranking.

**Table 2. Summary of Results from the Tappy Keystroke Dataset.**

| Feature Selection | Number of Features Selected (*from Ranked List) | SP | DI | EqO | Bacc |
|---|---|---|---|---|---|
| Regular | 19 | -0.0095 ± 0.1531 | 1.1846 ± 0.4026 | -0.0667 ± 0.2424 | 0.8262 ± 0.0578 |
| Fair | 19* | -0.0003 ± 0.0764 | 0.9822 ± 0.2127 | 0.0542 ± 0.0638 | 0.7637 ± 0.0268 |

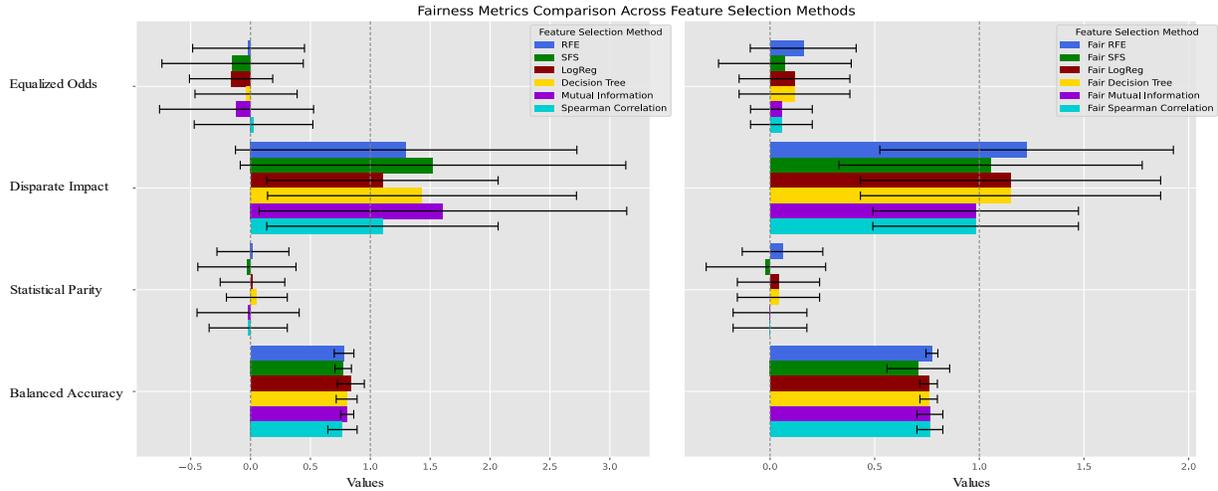

**Figure 3.** Comparison of fairness metrics (DI, SP and EqO) and Bacc for individual feature selection methods and their fair counterparts on the Tappy Keystroke Dataset.

*Clinical and Molecular Features data for Glioma Grading*

We selected the top-20 features from correlation based feature selection to align our fair feature selection approach for the analysis of the Glioma Grading dataset. Utilizing a 5-fold cross-validation method combined with 100 bootstrap sampling, we observed improvements in fairness metrics under the fairness-based feature selection: SP was enhanced to 0.0546 ± 0.0329 from its original value, DI was reduced to 1.1691 ± 0.0906, and EqO significantly enhanced to 0.0224 ± 0.0310, indicating improved fairness across these metrics. Additionally, Bacc experienced a notable increase to 0.8751 ± 0.0072. These results are summarized in Table 3 and graphically represented in Figure 4. A comparison of the individual features selection methods and their 'Fair' counterpart are illustrated in Figure 5.

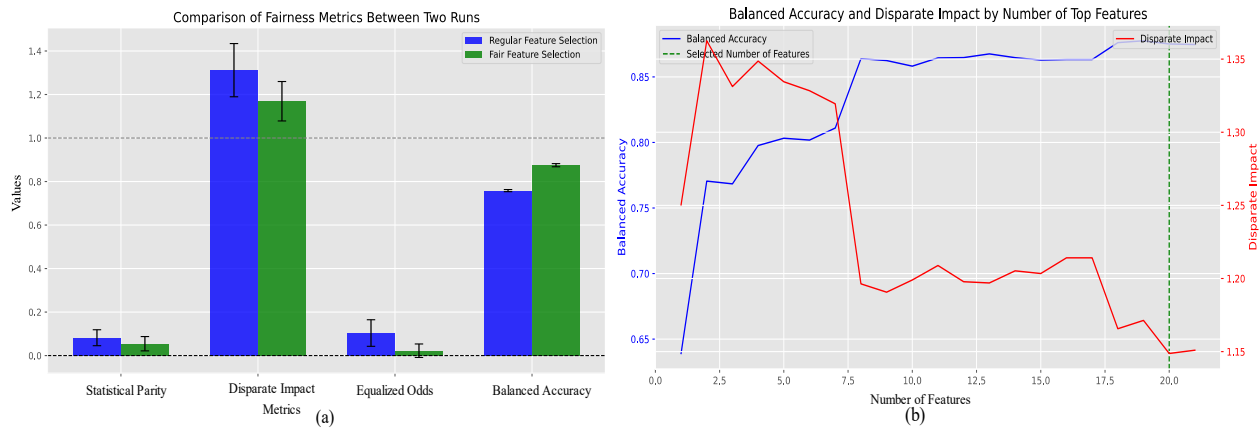

**Figure 4.** Clinical and Molecular Features Data for Glioma Grading analysis: (a) Contrast in fairness metrics (DI, SP and EqO) and Bacc between conventional feature selection and fairness-oriented feature selection approaches. (b) The relationship between the number of top features considered from the ranking and the changes in fairness metrics and accuracy.

**Table 3.** Summary of Results from the Clinical and Molecular Features data for Glioma Grading.

| Feature Selection | Number of Features Selected (*from Ranked List) | SP | DI | EqO | Bacc |
|---|---|---|---|---|---|
| Regular | 20 | 0.0821 ± 0.0367 | 1.3119 ± 0.1223 | 0.1037 ± 0.0611 | 0.7583 ± 0.0048 |
| Fair | 20* | 0.0546 ± 0.0329 | 1.1691 ± 0.0906 | 0.0224 ± 0.0310 | 0.8751 ± 0.0072 |

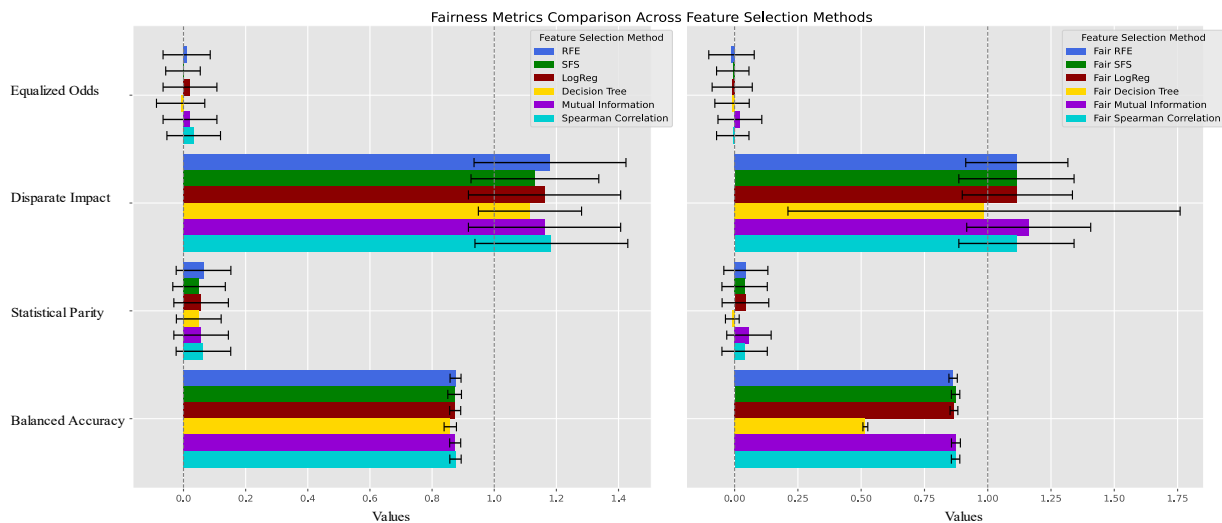

**Figure 5.** Comparison of fairness metrics (DI, SP and EqO) and Bacc for individual feature selection methods and their fair counterparts on the Clinical and Molecular Features Data for Glioma Grading.

*Hospital Admission Data for Coronary Artery Disease Hospital Admission Data for Coronary Artery Disease*
In the analysis of the Hospital Admission Data for Coronary Artery Disease, we adjusted the top features selected by correlation-based approach to match the 25 features selected by fair feature selection method. The correlation-based method yielded a Bacc of 0.6955 ± 0.0067. In contrast, the fairness-oriented approach led to an improved Bacc of 0.7099 ± 0.0078. Moreover, the fairness metrics demonstrated notable improvements: SP was enhanced from -0.1481 ± 0.0060 to -0.1108 ± 0.0109, and DI improved from 0.8180 ± 0.0066 to 0.8606 ± 0.0125. Additionally, EqO was notably reduced from -0.0699 ± 0.0115 to -0.0359 ± 0.0142, indicating better fairness. These results are summarized in Table 4 and illustrated in Figure 6. Figure 7 illustrates a contrast between the individual feature selection methods used in this dataset and their 'Fair' counterparts.

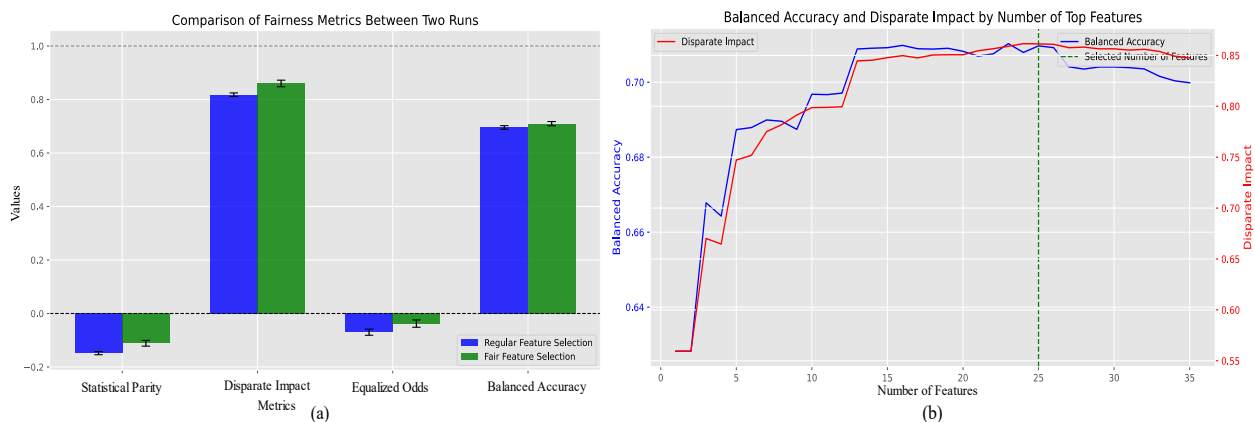

**Figure 6.** Hospital Admission Data for Coronary Artery Disease analysis: (a) Contrast in fairness metrics (DI, SP and EqO) and Bacc between conventional feature selection and fairness-oriented feature selection approaches. (b) Correlation between the number of top features selected based on ranking and the consequent changes in fairness metrics and accuracy.

**Table 4. Summary of Results from the Hospital Admission Data for Coronary Artery Disease.**

| Feature Selection | Number of Features Selected (*from Ranked List) | SP | DI | EqO | Bacc |
|---|---|---|---|---|---|
| Regular | 25 | -0.1481 ± 0.0060 | 0.8180 ± 0.0066 | -0.0699 ± 0.0115 | 0.6955 ± 0.0067 |
| Fair | 25* | -0.1108 ± 0.0109 | 0.8606 ± 0.0125 | -0.0359 ± 0.0142 | 0.7099 ± 0.0078 |

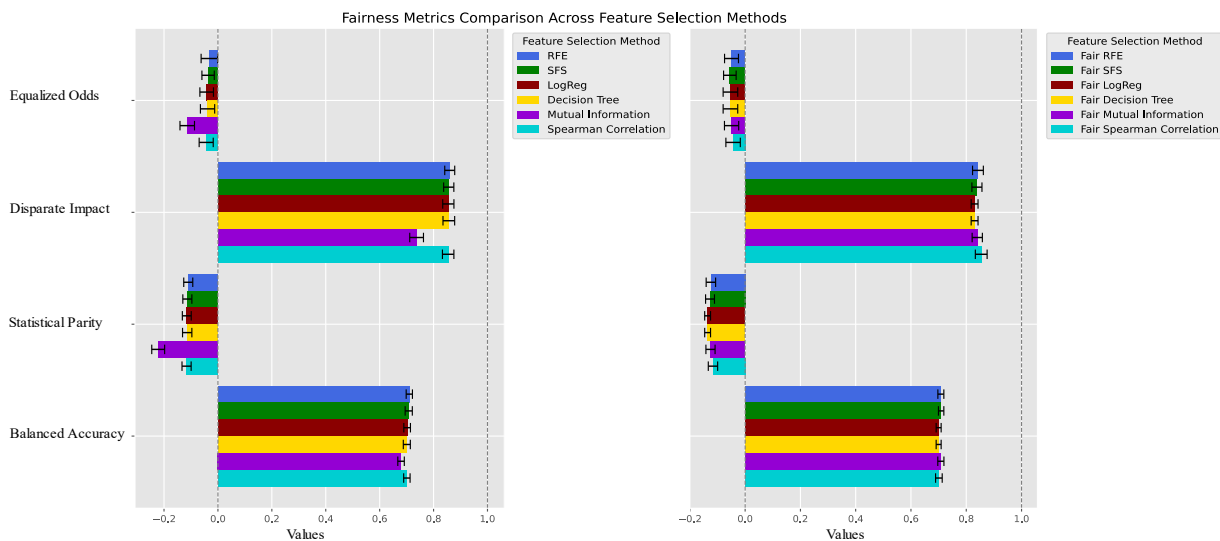

**Figure 7.** Comparison of fairness metrics (DI, SP and EqO) and Bacc for individual feature selection methods and their fair counterparts on the Clinical and Molecular Features Data for Hospital Admission Data.

**Discussion**

We investigated the implementation of a fair feature selection process to reduce biases in machine learning for healthcare. We evaluated this technique on gender-related biases within three distinct publicly available healthcare datasets: Tappy Keystroke (for Parkinson's detection), Clinical and Molecular Features data (for Glioma Grading), and Hospital Admission Data (for Coronary Artery Disease). The observed enhancements in fairness metrics, notably DI and SP, while maintaining Bacc, support the hypothesis that fairness and accuracy can be integrated successfully into predictive models through appropriate fair feature selection techniques.

Our work indicates that a generalized fair feature selection framework, considering both distributive and procedural fairness, is achievable. The results from our analysis show that combining multiple feature selection methods outperforms single method approaches in terms of fairness and accuracy. Using both supervised and information-centric measures, our framework offers a balanced improvement in fairness and accuracy for healthcare diagnostic models.

There are several notable limitations to our work that should be addressed in follow-up studies. First and foremost, we only evaluated our method on gender biases. We only studied two genders rather than considering a broader spectrum of gender identities, primarily due to the lack of this information in the publicly available datasets that we evaluated our method on. Our approach requires significant computational resources. Finally, the success of our bias mitigation strategy is reliant on the quality and comprehensiveness of the data; substandard data can reduce the effectiveness of bias correction via feature selection.

Future endeavors should aim to expand this evaluation to evaluate a wider range of demographic factors. Enhancements to the efficiency and fairness-awareness of our feature selection approach can also be made, making the technique more universally applicable and effective.

## Conclusion

We address bias in machine learning for healthcare by implementing a specialized fair feature selection strategy. By partitioning the datasets according to the sensitive attributes and utilizing an array of feature selection techniques separately on each partition, we are able to mitigate biases while improving the overall fairness of the models with minimal impact on balanced accuracy. This work contributes to the expanding field of ethical artificial intelligence in healthcare by presenting a method to ensure equity in predictive modeling—a critical consideration in healthcare settings where decision-making profoundly affects outcomes.

## Acknowledgements

Research reported in this publication was supported by the Office of the Director, National Institutes of Health Common Fund under award number 1OT2OD032581-01. The work is solely the responsibility of the authors and does not necessarily represent the official view of AIM-AHEAD or the National Institutes of Health. This work was supported by the AIM-AHEAD Coordinating Center, funded by the NIH. We used the generative AI tool ChatGPT by OpenAI to aid in editing the grammar of the manuscript.

**Name & Address of the Training Program**

MS in Computer Science, Department of Information and Computer Sciences, University of Hawaii at Manoa, 1680 East-West Road, Honolulu, HI 96822, USA.

**Author Contributions**

Data collection, methodology, analysis, review of literature, evaluation of results, writing (original manuscript preparation & editing) – MRSZ, Conceptualization, resources, funding acquisition, supervision, writing (reviewing manuscript) – PW. PW is the sole advisor of the student (MRSZ).

## References

1. Sharma K, Kaur A, Gujral S. Brain tumor detection based on machine learning algorithms. Int J Comput Appl. 2014 Oct 18;103(1):7–11.
2. Bazazeh D, Shubair R. Comparative study of machine learning algorithms for breast cancer detection and diagnosis. In: 2016 5th International Conference on Electronic Devices, Systems and Applications (ICEDSA) [Internet]. 2016 [cited 2024 Mar 7]. p. 1–4. Available from: https://ieeexplore.ieee.org/abstract/document/7818560
3. Kourou K, Exarchos TP, Exarchos KP, Karamouzis MV, Fotiadis DI. Machine learning applications in cancer prognosis and prediction. Comput Struct Biotechnol J. 2015 Jan 1;13:8–17.
4. Baghdadi NA, Farghaly Abdelaliem SM, Malki A, Gad I, Ewis A, Atlam E. Advanced machine learning techniques for cardiovascular disease early detection and diagnosis. J Big Data. 2023 Sep 17;10(1):144.
5. Cohn JN, Hoke L, Whitwam W, Sommers PA, Taylor AL, Duprez D, et al. Screening for early detection of cardiovascular disease in asymptomatic individuals. Am Heart J. 2003 Oct 1;146(4):679–85.
6. Bokolo BG, Liu Q. Deep learning-based depression detection from social media: comparative evaluation of ml and transformer techniques. Electronics. 2023 Jan;12(21):4396.
7. Applied Sciences | Free Full-Text | Individualized stress mobile sensing using self-supervised pre-training [Internet]. [cited 2024 Mar 7]. Available from: https://www.mdpi.com/2076-3417/13/21/12035
8. Rahat Shahriar Zawad Md, Yeaminul Haque Md, Kaiser MS, Mahmud M, Chen T. Computational intelligence in depression detection. In: Chen T, Carter J, Mahmud M, Khuman AS, editors. Artificial intelligence in healthcare: recent applications and developments [Internet]. Singapore: Springer Nature; 2022 [cited 2024 Mar 7]. p. 145–63. (Brain Informatics and Health). Available from: https://doi.org/10.1007/978-981-19-5272-2_7
9. Çubukçu HC, Topcu Dİ, Bayraktar N, Gülşen M, Sarı N, Arslan AH. Detection of covid-19 by machine learning using routine laboratory tests. Am J Clin Pathol. 2021 Nov 17;aqab187.
10. Rajaraman S, Candemir S, Xue Z, Alderson PO, Kohli M, Abuya J, et al. A novel stacked generalization of models for improved TB detection in chest radiographs. In: 2018 40th Annual International Conference of the IEEE Engineering in Medicine and Biology Society (EMBC) [Internet]. Honolulu, HI: IEEE; 2018 [cited 2024 Mar 7]. p. 718–21. Available from: https://ieeexplore.ieee.org/document/8512337/

11. Cirillo D, Catuara-Solarz S, Morey C, Guney E, Subirats L, Mellino S, et al. Sex and gender differences and biases in artificial intelligence for biomedicine and healthcare. Npj Digit Med. 2020 Jun 1;3(1):1–11.
12. Investigating for bias in healthcare algorithms: a sex-stratified analysis of supervised machine learning models in liver disease prediction - PMC [Internet]. [cited 2024 Mar 7]. Available from: https://www.ncbi.nlm.nih.gov/pmc/articles/PMC9039354/
13. Jain A, Brooks JR, Alford CC, Chang CS, Mueller NM, Umscheid CA, et al. Awareness of racial and ethnic bias and potential solutions to address bias with use of health care algorithms. JAMA Health Forum. 2023 Jun 2;4(6):e231197.
14. Adler NE, Glymour MM, Fielding J. Addressing social determinants of health and health inequalities. JAMA. 2016 Oct 25;316(16):1641–2.
15. Kamishima T, Akaho S, Asoh H, Sakuma J. Fairness-aware classifier with prejudice remover regularizer. In: Flach PA, De Bie T, Cristianini N, editors. Machine Learning and Knowledge Discovery in Databases. Berlin, Heidelberg: Springer; 2012. p. 35–50. (Lecture Notes in Computer Science).
16. Agarwal A, Beygelzimer A, Dudik M, Langford J, Wallach H. A reductions approach to fair classification. In: Proceedings of the 35th International Conference on Machine Learning [Internet]. PMLR; 2018 [cited 2024 Mar 7]. p. 60–9. Available from: https://proceedings.mlr.press/v80/agarwal18a.html
17. Goh G, Cotter A, Gupta M, Friedlander MP. Satisfying real-world goals with dataset constraints. In: Advances in Neural Information Processing Systems [Internet]. Curran Associates, Inc.; 2016 [cited 2024 Mar 7]. Available from: https://proceedings.neurips.cc/paper/2016/hash/dc4c44f624d600aa568390f1f1104aa0-Abstract.html
18. Salvador T, Cairns S, Voleti V, Marshall N, Oberman A. FairCal: fairness calibration for face verification [Internet]. arXiv; 2022 [cited 2024 Mar 12]. Available from: http://arxiv.org/abs/2106.03761
19. Karimi-Haghighi M, Hernández-Leo D, Castillo C, Oliver VM. Predicting early dropout: calibration and algorithmic fairness considerations.
20. Yang K, Huang B, Stoyanovich J, Schelter S. Fairness-aware instrumentation of preprocessing pipelines for machine learning. 2020;
21. Biswas S, Rajan H. Fair preprocessing: towards understanding compositional fairness of data transformers in machine learning pipeline. In: Proceedings of the 29th ACM Joint Meeting on European Software Engineering Conference and Symposium on the Foundations of Software Engineering [Internet]. New York, NY, USA: Association for Computing Machinery; 2021 [cited 2024 Mar 12]. p. 981–93. (ESEC/FSE 2021). Available from: https://dl.acm.org/doi/10.1145/3468264.3468536
22. Celis LE, Keswani V, Vishnoi N. Data preprocessing to mitigate bias: a maximum entropy based approach. In: Proceedings of the 37th International Conference on Machine Learning [Internet]. PMLR; 2020 [cited 2024 Mar 12]. p. 1349–59. Available from: https://proceedings.mlr.press/v119/celis20a.html
23. Rathore S. Model agnostic feature selection for fairness [Internet]. [Kingston, RI]: University of Rhode Island; 2022 [cited 2024 Mar 7]. Available from: https://digitalcommons.uri.edu/theses/2291
24. Galhotra S, Shanmugam K, Sattigeri P, Varshney KR. Causal feature selection for algorithmic fairness. In: Proceedings of the 2022 International Conference on Management of Data [Internet]. Philadelphia PA USA: ACM; 2022 [cited 2024 Mar 7]. p. 276–85. Available from: https://dl.acm.org/doi/10.1145/3514221.3517909
25. Khodadadian S, Nafea M, Ghassami A, Kiyavash N. Information theoretic measures for fairness-aware feature selection [Internet]. arXiv; 2021 [cited 2024 Mar 7]. Available from: http://arxiv.org/abs/2106.00772
26. Xing X, Liu H, Chen C, Li J. Fairness-aware unsupervised feature selection [Internet]. arXiv; 2021 [cited 2024 Mar 7]. Available from: http://arxiv.org/abs/2106.02216
27. Grgic-Hlacˇa N, Zafar MB, Gummadi KP, Weller A. The case for process fairness in learning: feature selection for fair decision making.
28. Grgić-Hlača N, Zafar MB, Gummadi KP, Weller A. Beyond distributive fairness in algorithmic decision making: feature selection for procedurally fair learning. Proc AAAI Conf Artif Intell [Internet]. 2018 Apr 25 [cited 2024 Mar 7];32(1). Available from: https://ojs.aaai.org/index.php/AAAI/article/view/11296
29. Belitz C, Jiang L, Bosch N. Automating procedurally fair feature selection in machine learning. In: Proceedings of the 2021 AAAI/ACM Conference on AI, Ethics, and Society [Internet]. New York, NY, USA: Association for Computing Machinery; 2021 [cited 2024 Mar 14]. p. 379–89. (AIES '21). Available from: https://doi.org/10.1145/3461702.3462585
30. Brookhouse J, Freitas A. Fair Feature Selection: A comparison of multi-objective genetic algorithms [Internet]. arXiv; 2023 [cited 2024 Mar 14]. Available from: http://arxiv.org/abs/2310.02752
31. Ayaz Z, Naz S, Khan NH, Razzak I, Imran M. Automated methods for diagnosis of Parkinson's disease and predicting severity level. Neural Comput Appl. 2023 Jul 1;35(20):14499–534.

32. Zhao T, Dai E, Shu K, Wang S. Towards fair classifiers without sensitive attributes: exploring biases in related features. In: Proceedings of the Fifteenth ACM International Conference on Web Search and Data Mining [Internet]. 2022 [cited 2024 Mar 14]. p. 1433–42. Available from: http://arxiv.org/abs/2104.14537
33. Adams WR. High-accuracy detection of early Parkinson's Disease using multiple characteristics of finger movement while typing. PLOS ONE. 2017 Nov 30;12(11):e0188226.
34. Goldberger AL, Amaral LA, Glass L, Hausdorff JM, Ivanov PC, Mark RG, et al. PhysioBank, PhysioToolkit, and PhysioNet: components of a new research resource for complex physiologic signals. Circulation. 2000 Jun 13;101(23):E215-220.
35. Tasci E, Zhuge Y, Kaur H, Camphausen K, Krauze AV. Hierarchical voting-based feature selection and ensemble learning model scheme for glioma grading with clinical and molecular characteristics. Int J Mol Sci. 2022 Nov 16;23(22):14155.
36. Coronary Artery Disease - Analysis [Internet]. [cited 2024 Mar 7]. Available from: https://www.kaggle.com/datasets/homelysmile/datacad
37. Bollepalli SC, Sahani AK, Aslam N, Mohan B, Kulkarni K, Goyal A, et al. An optimized machine learning model accurately predicts in-hospital outcomes at admission to a cardiac unit. Diagnostics. 2022 Jan 19;12(2):241.
38. Gordon AD, Breiman L, Friedman JH, Olshen RA, Stone CJ. Classification and regression trees. Biometrics. 1984 Sep;40(3):874.
39. Applied logistic regression, 3rd Edition | Wiley [Internet]. [cited 2024 Mar 7]. Available from: https://www.wiley.com/en-us/Applied+Logistic+Regression%2C+3rd+Edition-p-9780470582473
40. Elements of information theory | Wiley Online Books [Internet]. [cited 2024 Mar 7]. Available from: https://onlinelibrary.wiley.com/doi/book/10.1002/047174882X
41. Spearman correlation coefficients, differences between - Myers - Major Reference Works - Wiley Online Library [Internet]. [cited 2024 Mar 7]. Available from: https://onlinelibrary.wiley.com/doi/10.1002/0471667196.ess5050.pub2
42. Kohavi R. A study of cross-validation and bootstrap for accuracy estimation and model selection. In 1995 [cited 2024 Mar 7]. Available from: https://www.semanticscholar.org/paper/A-Study-of-Cross-Validation-and-Bootstrap-for-and-Kohavi/8c70a0a39a686bf80b76cb1b77f9eef156f6432d
43. The balanced accuracy and its posterior distribution | IEEE Conference Publication | IEEE Xplore [Internet]. [cited 2024 Mar 7]. Available from: https://ieeexplore.ieee.org/document/5597285
44. Feldman M, Friedler S, Moeller J, Scheidegger C, Venkatasubramanian S. Certifying and removing disparate impact [Internet]. arXiv; 2015 [cited 2024 Mar 7]. Available from: http://arxiv.org/abs/1412.3756
45. Dwork C, Hardt M, Pitassi T, Reingold O, Zemel R. Fairness through awareness [Internet]. arXiv; 2011 [cited 2024 Mar 7]. Available from: http://arxiv.org/abs/1104.3913
46. Hardt M, Price E, Srebro N. Equality of opportunity in supervised learning [Internet]. arXiv; 2016 [cited 2024 Mar 7]. Available from: http://arxiv.org/abs/1610.02413